\documentclass[letterpaper]{article} 
\usepackage{times}  
\usepackage{helvet}  
\usepackage{courier}  
\usepackage[hyphens]{url}  
\usepackage{graphicx} 
\usepackage{natbib}
\setcitestyle{numbers,square}

\usepackage{caption} 
\usepackage[final]{neurips_2023}
\usepackage{times}
\usepackage{soul}
\usepackage{url}
\usepackage[utf8]{inputenc}
\usepackage{amsmath}
\usepackage{amsthm}
\usepackage{booktabs}
\usepackage{algorithm}
\usepackage{algorithmic}
\usepackage[switch]{lineno}
\usepackage{textcomp}
\usepackage[table]{xcolor}
\usepackage{multirow}
\usepackage{url}
\usepackage{amsfonts}
\usepackage{pifont}
\usepackage{caption}
\usepackage{subcaption}

\usepackage{newfloat}
\usepackage{listings}
\DeclareCaptionStyle{ruled}{labelfont=normalfont,labelsep=colon,strut=off} 
\floatstyle{ruled}
\newfloat{listing}{tb}{lst}{}
\floatname{listing}{Listing}
\newcommand{\cmark}{\ding{51}}%
\author{
  Tanya Akumu \\
  IBM Research Africa\\ 
  Nairobi, Kenya\\
  \And
  Celia Cintas\\
  IBM Research Africa \\ 
  Nairobi, Kenya\\
  \AND
  Girmaw Abebe Tadesse \\
  Microsoft AI for Good Research Lab~\footnotemark{}\\
  Nairobi, Kenya \\
  \And
  Adebayo Oshingbesan \\
  IBM Research Africa \\
  Nairobi, Kenya  \\
  \And
  Skyler Speakman \\
  IBM Research Africa \\ 
  Nairobi, Kenya \\
  \And
  Edward McFowland III \\
  Harvard Business Institute \\ 
  Masachusetts, USA \\
}

\title{Efficient Representation of the Activation Space in Deep Neural Networks}

\begin{document}

\maketitle

\begin{abstract}

The representations of the activation space of deep neural networks (DNNs) are widely utilized for tasks like natural language processing, anomaly detection and speech recognition. Due to the diverse nature of these tasks and the large size of DNNs, an efficient and task-independent representation of activations becomes crucial. Empirical $p$-values have been used to quantify the relative strength of an observed node activation compared to activations created by already-known inputs.   Nonetheless, keeping raw data for these calculations increases memory resource consumption and raises privacy concerns. To this end, we propose a model-agnostic framework for creating representations of activations in DNNs using node-specific histograms to compute $p$-values of observed activations without retaining already-known inputs. Our proposed approach demonstrates promising potential when validated with multiple network architectures across various downstream tasks and compared with the kernel density estimates and brute-force empirical baselines. In addition, the framework reduces memory usage by 30\% with up to 4 times faster $p$-value computing time while maintaining state-of-the-art detection power in downstream tasks such as the detection of adversarial attacks and synthesized content. Moreover, as we do not persist raw data at inference time, we could potentially reduce susceptibility to attacks and privacy issues.
\footnotetext[1]{This work was done while he was at IBM Research Africa.}

\end{abstract}

\section*{Introduction}

The ubiquity of applications that utilize deep neural networks (DNNs) has brought the emergence of a variety of architectures and improvements in the accuracy of deep learning tasks such as natural language processing, signal, and image processing~\cite{touvron2023llama,huang2022fastdiff,ramesh2022hierarchical}. These DNNs learn representations, which are abstractions of the data in the shape of feature vectors, that capture the most relevant characteristics for the task at hand~\cite{bengio2013representation}. Investigating the representations learned by DNNs is crucial for enhancing algorithms to improve performance on downstream tasks, model interpretability, fairness, and robust explanation~\cite{samek2016evaluating, montavon2018methods, samek2017explainable, liang2022advances, li2022interpretable}. However, working directly on the activations derived from previous inputs pose a significant challenge in terms of memory resource consumption and privacy concerns. Thus, efficient representations of activations, without substantial memory or runtime overhead, is critical to enable practical deployment in secure low-resource settings.

Existing works employed different strategies to derive these efficient representations by considering them as data-generating systems applying pattern detection techniques in the activation space of hidden layers. These patterns provide insights into the behavior of the DNN, and its performance on different tasks such as DNNs testing and interpretability~\cite{ma2018deepgauge,pei2017deepxplore}, fake content detection~\cite{ijcai2020p476}, and anomalous pattern detection ~\cite{cintas2022pattern}. Some of these approaches use empirical $p$-values to quantify the strength of observed node activations compared to activations created by already-known background for anomalous patterndetection~\cite{akinwande2020identifying,kim2022out,cintas2022towards}. 
This strategy cannot distinguish between tied likelihoods and lower likelihoods, introducing a bias towards higher $p$-values. Other approaches build representations with
Bayesian-based uncertainty estimation was also employed, followed with a kernel density estimation in the activation space for adversarial attack detection~\cite{feinman2017detecting}. 

Although these strategies have proven to be useful, we observe common challenges across the existing approaches. For instance, they require significant memory resources to handle high-dimensional inputs, and their applicability is often limited to specific tasks, choice of network architectures, and layers hindering their generalizability. Furthermore, retaining all the information of the activation space even at inference time poses privacy concerns and vulnerability to adversarial attacks~\cite{akhtar2018threat,li2022blacklight}. Consequently, characterizing the activation space of DNNs in a more efficient and task-independent manner is imperative.

This work presents a framework for creating simple but effective representations of activations in DNNs, which applies to any off-the-shelf pre-trained neural network in a way that optimizes memory usage and run-time. Further, it is designed to keep the minimal information about the activation space as possible by using node-specific histograms.
In summary, this paper makes the following contributions.
\begin{itemize}
    \item We propose a task-independent and model-agnostic technique for obtaining representation of activations in DNNs.
    \item We employ node-specific histograms that minimizes the amount of information needed to represent the activations, thereby achieving effective use of memory usage and potentially reducing vulnerability of privacy and adversarial concerns. 
    \item We compare our approach with two task-independent representation characterization baselines and show empirically that we can obtain up to 30\% memory savings while maintaining state-of-the-art performances in various downstream tasks~\cite{akinwande2020identifying,cintas2022towards,you2018graph}.
    \item We reduce the run-time of computing $p$-value ranges by up to 4 times compared to existing methods. By employing $p$-value ranges instead of traditional empirical $p$-values for downstream tasks, we may enhance the robustness of the representations.
    \item We validate our proposed approach using several pre-trained models of different architectures (Generative Adversarial Networks, Graph models, and Convolutional Neural Networks), and multiple datasets (WikiArt~\cite{artgan2018}, ZINC~\cite{irwin2005zinc,sterling2015zinc}, MNIST~\cite{lecun1998mnist}). 
\end{itemize}

\section*{Related Work} \label{related_work}
In the field of deep learning, the activation space of neural networks plays a crucial role in determining their performance on various tasks. To better understand this space, researchers have developed several methods for characterizing the activation patterns of DNNs. In this section, we discuss two categories of these methods: task-dependent characterization and task-agnostic characterization of the activation space of DNNs. A summary of the comparison of the key characteristics of these techniques is given in Table~\ref{tab:rl_summary}.

\subsection*{Task-dependent 
characterization methods}

~\cite{feinman2017detecting}
utilized the Bayesian-encoded uncertainty in dropout neural networks and KDE on the subspace learned by the model for adversarial attack detection. In another stream of work, testing and verification of DNNs were performed with node coverage criteria. For example, DeepXplore~\cite{pei2017deepxplore} utilizes a pre-defined threshold $\tau$ and computes the percentage of node activations greater than  $\tau$. DeepGauge~\cite{ma2018deepgauge} extends this criterion to consider the output value of each neuron by using k-multi-section node coverage, which divides the output range of each node activations into $k$ chunks. Similarly, FakeSpotter~\cite{ijcai2020p476} monitors the node behavior of a DNN using a threshold calculated from the average values of activations in each layer, driven from training samples. These studies mainly focus on creating representations that apply only to their specific downstream task and model architectures, rendering their approach non-generalizable.

\begin{table*}[!ht]

\caption{Summarization of existing methods in the state-of-the-art representation characterization in deep neural networks compared to our proposed approach.
}
\centering
\resizebox{\linewidth}{!}{
\begin{tabular}{@{}llcccc@{}}
 
\toprule

\ Approach & Description & Activation representation & Compressed inputs  & Model-agnostic & Multi-task 
 \\\midrule
 \cite{ijcai2020p476} & Average-based thresholding &  \cmark  &  &  & 

\\
\cite{feinman2017detecting} & Density estimates & \cmark &  &  \cmark & \cmark   

\\
 \cite{cintas2022pattern}  & Empirical $p$-values & \cmark &  & \cmark &\cmark 
\\
\midrule
 Proposed  & Node-specific histograms & \cmark & \cmark & \cmark &\cmark 
\\
\bottomrule
\end{tabular}
}
\label{tab:rl_summary}
\end{table*}
 
\subsection*{Task-agnostic characterization methods}
To overcome the limitations of task-dependent representation characterization, most recent works ~\cite{cintas2022pattern,akinwande2020identifying,kim2022out,cintas2022towards,kim2022spatially} have employed empirical $p$-value based strategies to characterize the representations of node activations in a DNN. These works utilize subset scanning~\cite{mcfowland-fgss-2013}, which identifies subsets of neurons with activation patterns that deviate from the expected distribution. The empirical $p$-values are calculated to determine the likelihood of observing such deviations by chance. These studies use empirical $p$-value based representations to understand the features detected by the network and their relationships without being limited to specific downstream tasks.

Although these techniques can be task-independent, they still require all the information about the activation space to be known. This makes them memory inefficient, slower to use at inference time, and more susceptible to privacy concerns and adversarial attacks.

In this work, we propose a representation method that is agnostic to downstream tasks and network architectures. Furthermore, our method compresses the input required, therefore reducing the memory used to perform a downstream task. This makes it faster and possibly less susceptible to adversarial and privacy issues. 

\section*{Preliminaries}
\label{sec:preliminaries}
In characterizing node activations of a neural network, subset scanning~\cite{mcfowland-fgss-2013} uses $p$-values to assess the statistical significance of activation pattern deviations in intermediate layers and to identify subsets of deviating neurons. This section outlines current $p$-value calculation methods.

\subsubsection*{Empirical $p$-values estimation}
To describe the estimation of empirical $p$-values, let $A^{\mathcal{B}}_{lj}$ be the matrix of activations in a node $j\in O_l$ of $l \in \mathcal{L}$ layer of a DNN. These latent vectors are drawn from $\mathcal{B}$  background samples containing $\mathcal{N_B}$ samples, at each of $O_l$ nodes in a pre-trained neural network.  Let $A^{\mathcal{T}_k}_{lj}$ be the matrix of activations induced by $\mathcal{L}$ latent vectors drawn from the test set, ${\mathcal{T}_k}$. We compute an empirical $p$-value for each $A^{\mathcal{T}_k}_{lj}$, as a measurement for how anomalous the activation value of  each $k^{th}$ sample in the test set, $\mathcal{T}_k$ , is at node $j \in O_l $, following~\cite{cintas2020detecting,akinwande2020identifying}. 
This $p$-value, $p^k_{lj}$, is the proportion of activations from the $\mathcal{B}$ background samples, $A^{\mathcal{B}}_{lj}$, that are larger or equal to the activation from an evaluation sample $A^{\mathcal{T}_k}_{lj}$ at node $j \in O_l $. The empirical $p$-values are calculated as in Equation~\ref{empirical_pvalue}. This method treats tied likelihoods the same as lower likelihood values, thus introducing a bias towards higher $p$-values when ties are present. Our approach extends these traditional empirical $p$-values to $p$-value ranges which accounts for ties in the likelihoods, plausible making more robust representations. 

\begin{equation}
    p^k_{lj} = \frac{1+\sum_{l=1}^{|\mathcal{L}|} I(A^{\mathcal{B}}_{lj} \geq A^{\mathcal{T}_k}_{lj} )}{|\mathcal{N_B}|+1}
    \label{empirical_pvalue}
\end{equation}

where $I(\cdot)$ is the indicator function. A shift is added to the numerator and denominator so that a test activation that is larger than all activations from the background at that node is given a non-zero $p$-value.  

\subsubsection{Kernel Density Estimate $p$-values} 
Another technique for computing $p$-value based representation is using kernel density estimates (KDE)~\cite{feinman2017detecting,silverman1981using}. This technique estimates the unknown probability density function (PDF) of a given sample~\cite{parzen1962estimation} and computes $p$-values for each $k^{th}$ sample in the test set, $\mathcal{T}_k$,  as shown in Equation~\ref{eq:kde_pval}.

\begin{equation}
\label{eq:kde_pval}
     p^k_{lj} =
    \begin{cases}\int_{A^{\mathcal{T}_k}_{lj}}^{\Omega_j} \phi_{lj}\,dk & \text{for } A^{\mathcal{T}_k}_{lj} < \Omega_j,\\ \vspace{0.25pt}
    \frac{1}{ \mathcal{N_B} + 1}  ,&  \text{for } A^{\mathcal{T}_k}_{lj} \geq \Omega_j\\
    \end{cases}
\end{equation}

where $\Omega_j$ is the maximum observed activation in the background data at node $j \in O_l$, $\phi_{lj}$ is the PDF and $\mathcal{N_B}$ is the total number of samples in the background data. 

\section*{Proposed Approach}\label{framework}
\begin{figure}
    \centering
    \begin{subfigure}[b]{0.79\textwidth}
        \centering
        \includegraphics[width=\textwidth]{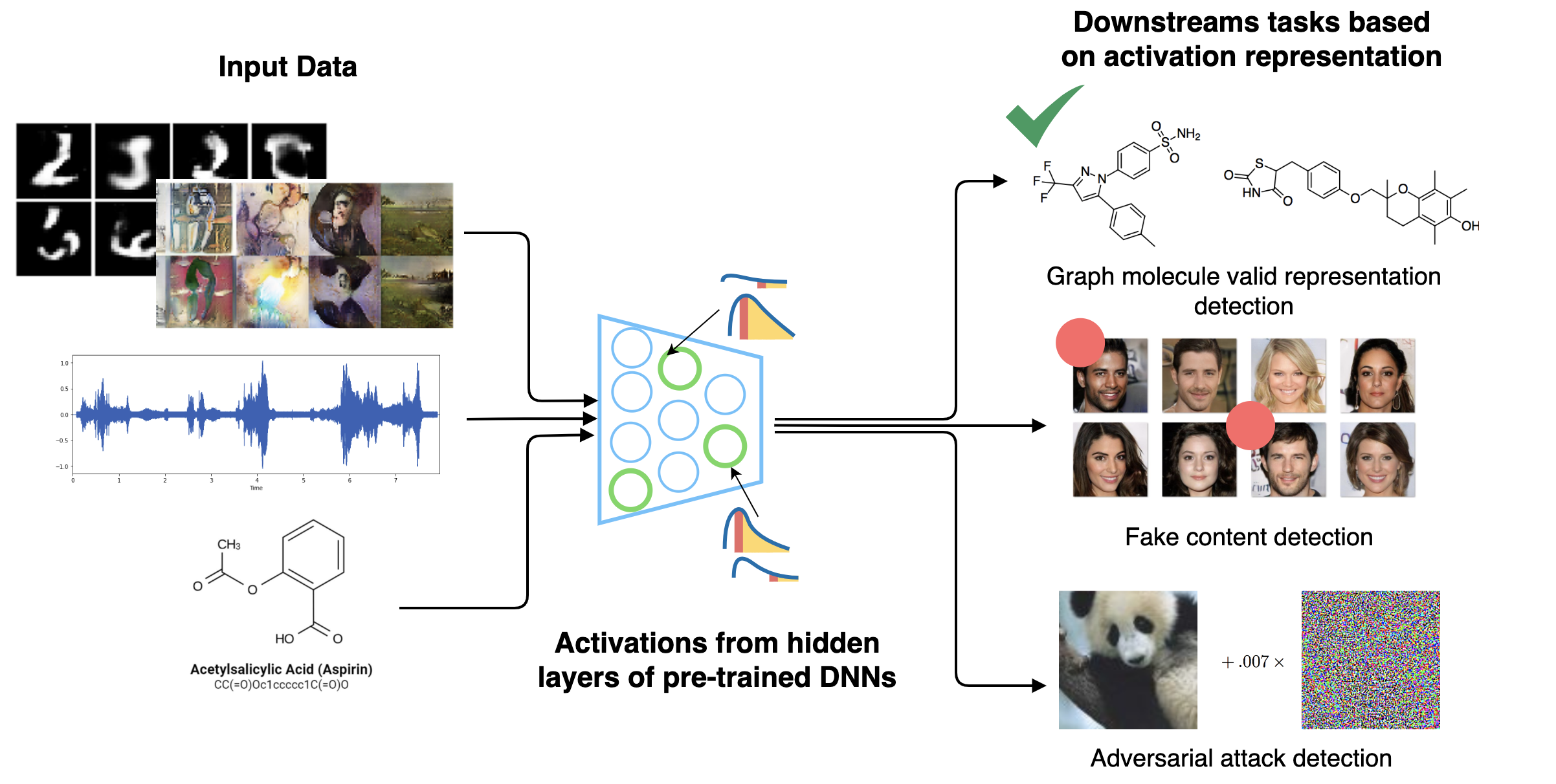}
    \subcaption{}
    \end{subfigure}
    
    \hfill
    \begin{subfigure}[b]{0.99\textwidth}
        \centering
        \includegraphics[width=\textwidth]{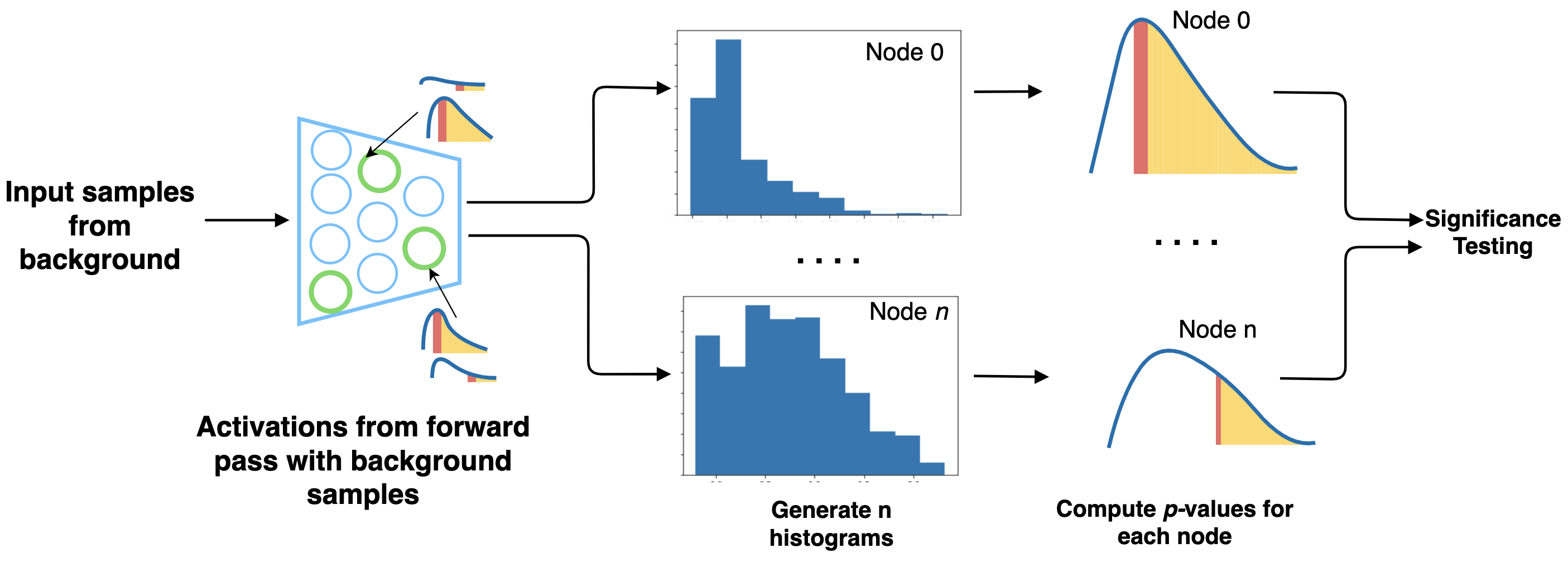}\subcaption{}
    \end{subfigure}
    \caption
    {Overview of our proposed framework for activation representation of DNNs. \textbf{(a)} First, we extract background activations from the hidden layers of a pre-trained model for three downstream tasks~\cite{cintas2022towards,you2018graph,akinwande2020identifying}. \textbf{(b)} Second, we create node-specific histograms consisting of bins and counts for each node in the network and use them to compute the $p$-values ranges of activations to be evaluated. Lastly, we use the $KS$-test to evaluate the $p$-value distributions.}
    \label{fig:flowchart}
\end{figure}

In this section, we describe the problem formulation, and the details of the proposed framework (see Fig.~\ref{fig:flowchart}).
The core elements of our method are \textit{Node-specific histograms}, \textit{$p$-values Ranges Computation}, and \textit{Significance testing}. 

Given a pre-trained DNN $M$, consisting of a set of intermediate layers, $\mathcal{L}$,  with each layer $l \in \mathcal{L}$ consisting of $O_l$ nodes, we aim to characterize the activation space of $M$ efficiently and in a generalizable way. We propose a method that employs node-specific histograms, $\Theta_{lj}$, to represent the activation space  $A^{\mathcal{B}}$ of the pre-trained model $M$ via efficiently driven $p$-values.
 
There are three steps to generate the new $p$-value representation, $p^k$, of the activations drawn from the nodes of the layers of model $M$.
First, $M$ is trained on a set of background samples, $\mathcal{B}$, containing $\mathcal{N_B}$ samples. The corresponding activations, $A^{\mathcal{B}}_{lj}$, are drawn at each $j \in O_l $ node of the $l\in \mathcal{L}$ layer in $M$. We create node-specific histograms, $\Theta_{lj}$ in the form of bin edges and counts,  for each node in the background activations, $A^{\mathcal{B}}_{lj}$. Second, given a set of test samples, $\mathcal{T}$, containing $N_\mathcal{T}$ samples, we extract test activations,  $A^{\mathcal{T}_k}_{lj}$, from the pre-trained model passed with the test samples. This is done for each $k^{th}$ sample in  $\mathcal{T}$, that will be utilized to compute a $p$-values $p^k_{lj}$, using the histogram bins edges and counts in $\Theta_{lj}$ and the test activations $A^{\mathcal{T}_k}_{lj}$. Finally, these $p$-values are  utilized for downstream tasks.  

\paragraph{Node-specific histograms} 
Once the activations drawn at each $j \in O_l $ node of the $l\in \mathcal{L}$ layer in $M$ are obtained, we create node-specific histograms for each node. We implement a histogram,$\Theta_{lj}$, for a node $j \in O_l$  described by the number of bins, $b$, and the height of the bins. This height represents the count of activations that fall into a specific bin. Since modal activations can cause spikes that will affect the number of bins, we only build histograms for activations outside the mode and assign a unique bin to represent modal activations. For activations outside the modal, we assign the number of bins as a maximum of the Freedman Diaconis estimator~\cite{freedman1981histogram}, and the Sturges estimator ~\cite{scott2009sturges} which computes the optimal bin width and consequently, the number of bins per node. We choose these estimators because they  take into account the data size and variability in the data. The Freedman Diaconis bin-width estimation is given by:

\begin{equation}
    h = 2 \frac{IQR}{\sqrt[3]{n}}
\end{equation}

where IQR is the interquartile range, and $n$ is the size of the data.

The number of bins estimated by Sturges method is given by:
\begin{equation}
    n_h = \log_2(n) + 1
\end{equation}

As a result, the bin-width of the histograms is non-uniform. We describe these histograms in the form of bin edges which tell us the number of bins as well as the boundaries for each bin and the height/count of each bin. For a test activation $A^{\mathcal{T}_k}_{lj}$, we find the bin to which it belongs by searching the bin edges of the node's histogram. Then we use the height of the obtained bin to compute its $p$-value as described in the next paragraph. 

\paragraph{$\mathbf{p}$-values ranges computation} 
In the Preliminaries section, we saw that traditional empirical $p$-values do not take into account ties in the likelihood leading to a bias for larger $p$-values. Our approach defines the $p$-value as a range between a maximum and a minimum to account for these ties. These ranges are made wider by using histogram summaries of the background distributions, likely making the representations more robust. We compute $p$-values for each activation $A^{\mathcal{T}_k}_{lj}$, as a measure of how anomalous the activation value at a node $j\in O_l$ of $l \in \mathcal{L}$ layer is. This $p$-value, $p^k_{lj}$, is the proportion of activations from the background input, $A^{\mathcal{B}}_{lj}$, that are greater than or equal to the observed activation, $A^{\mathcal{T}_k}_{lj}$. Given a histogram distribution, $\Theta_{lj}$ for a given node, we obtain the $p$-values using the height of the bins from the histogram (counts). This means that the $p$-values will correspond to their respective bins, and thus test activations falling into the same bin will have the same $p$-value.  Formally, our $p-$value ranges computation is defined as follows:

\begin{equation}
\begin{aligned}
\label{p-value-eq}
p^k_{lj} = [P_{min}(p^k_{lj}), P_{max}(p^k_{lj})]
\end{aligned}
\end{equation}

where:
\begin{equation}
    \label{p_min}
    P_{min}(p^k_{lj}) = \frac{\mathcal{N_B} - \sum_{i=b_{kj+1}}^{b_{nj}}{C_{ij}}}{\mathcal{N_B} + 1} 
\end{equation}

and,
\begin{equation}
    \label{p_max}
    P_{max}(p^k_{lj}) = \frac{\mathcal{N_B} - \sum_{i=b_{kj}}^{b_{nj}}{C_{ij} + 1}}{\mathcal{N_B} + 1} 
\end{equation}

where $b_{kj}$ represents the bin of the $k^{th}$ test activation of node $j \in O_l $, $b_{nj}$ is the $n^{th}$ bin of the histogram of node $j \in O_l $ with $n$ bins and $C_{ij}$ is the count/height of the $i^{th}$ bin.

\paragraph{Comparing learned representations} We compare the distribution of node-specific histogram $p$-values representation with the empirical $p$-values representation using goodness-of-fit statistics. Specifically, we use the two-sample Kolmogorov-Smirnov test~\cite{massey1951kolmogorov} to ascertain that the distributions of the empirical and node-specific histograms $p$-value representations are similar. To compare the $p$-value distributions of the baselines, we  draw a $p$-value uniformly at random from each range $[P_{min}(p^k_{lj}), P_{max}(p^k_{lj})]$ and apply the test to the drawn values. 

\begin{figure}[!ht]
    \centering
    \begin{tabular}{@{}lcc@{}}
    \toprule
    Model Architecture (Task) & Detection Power & Subset Scores Distribution\\
    \midrule
    \raisebox{60pt}{\parbox{4cm}{ArtGAN \\(Creative artifacts)}} & {\includegraphics[scale=0.45]{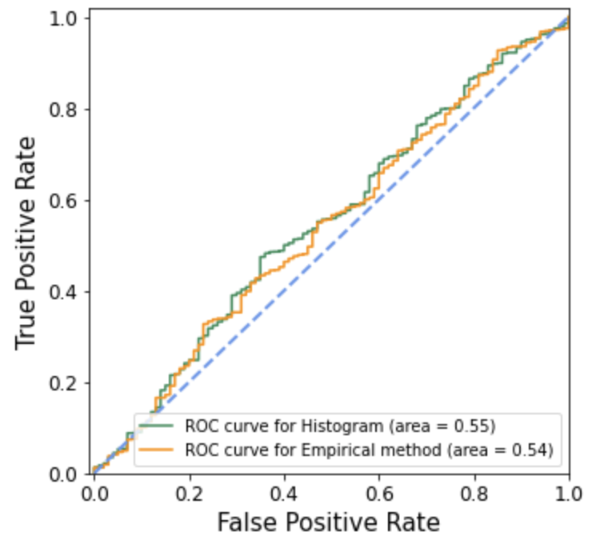}}  & {\includegraphics[scale=0.45]{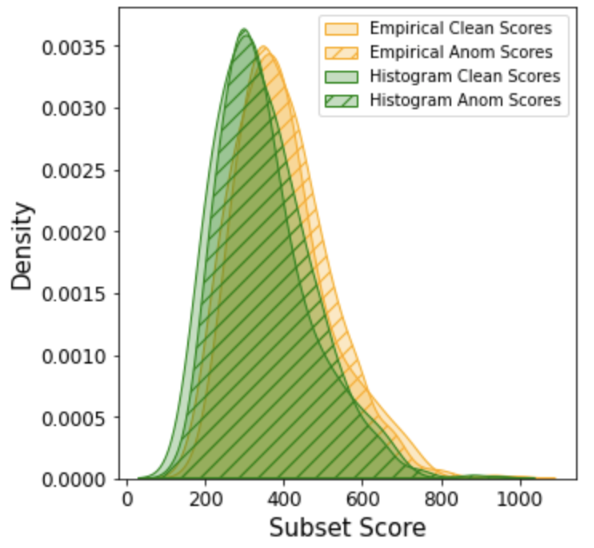}}\\
    \midrule
    \raisebox{60pt}{\parbox{4cm}{GCPN \\(Invalid graph molecules)}} & {\includegraphics[scale=0.45]{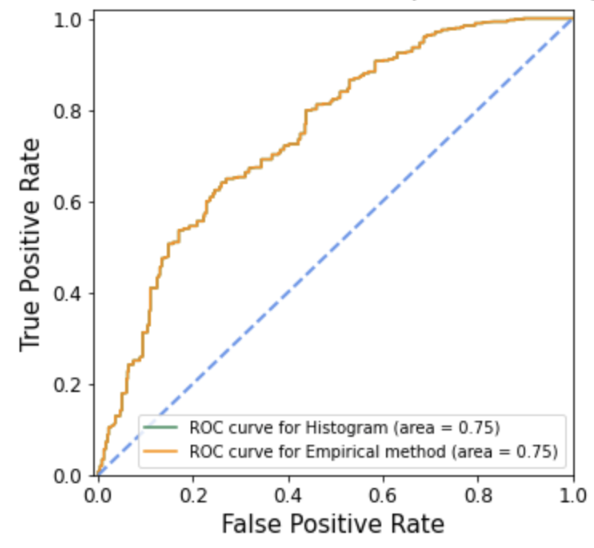}}  & {\includegraphics[scale=0.45]{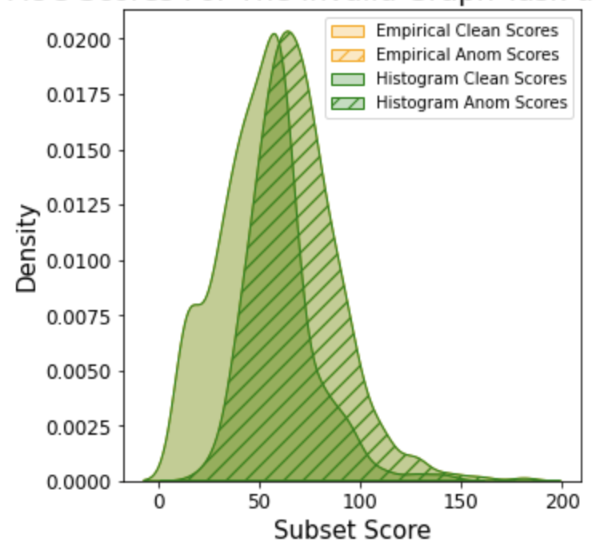}}\\
    \midrule
    \raisebox{60pt}{\parbox{4cm}{CNN \\(Adversarial attacks)}} & {\includegraphics[scale=0.45]{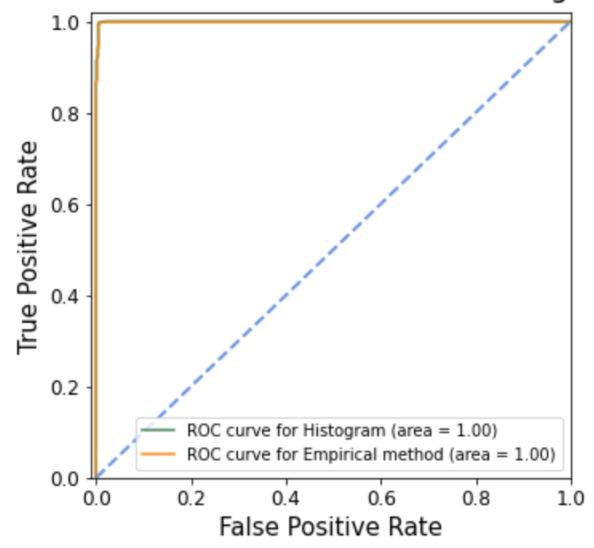}}  & {\includegraphics[scale=0.45]{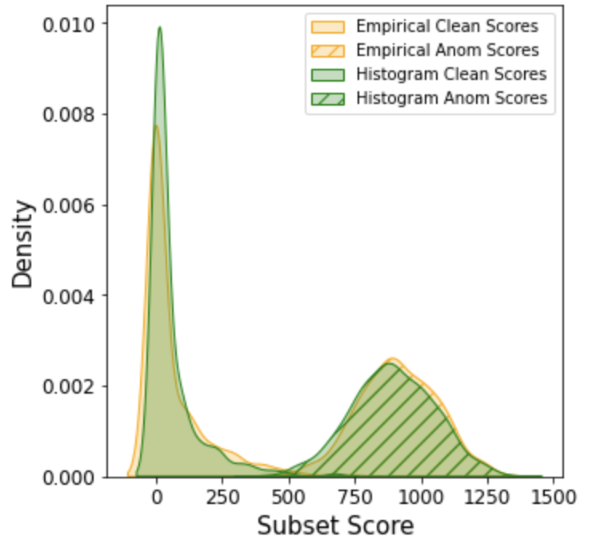}}\\
    \bottomrule
    \end{tabular}
    \caption{Comparison of Detection Power for three different downstream tasks~\cite{cintas2022towards,you2018graph,akinwande2020identifying} using our proposed node-specific histogram $p$-value representation and the Empirical $p$-values baseline~\cite{mcfowland-fgss-2013,speakman2018subset,feng-npss_graph-2014}.
    }
    \label{fig:detection_pow_hist}
\end{figure}

\begin{table}[!ht]
\caption{Memory usage measured in megabytes (MB) and learning representation run-time measured in seconds (s) of the different baselines for the ArtGAN using the ArtWiki Dataset~\cite{artgan2018}. $(^\ast)$Note that the empirical baseline does not learn any representation, but rather does a single sort of background activations when comparing the incoming test activations.
}
\centering
\begin{tabular}{@{}llll!{\vline}lll@{}}
\hline
 & \multicolumn{3}{c|}{Memory Usage (MB)} & \multicolumn{3}{c}{Learning representation run-time (s) $\pm$std} \\
 \hline
 \# Nodes ($n$)& Empirical & Proposed &KDE  &    Empirical$(^\ast)$   &   Proposed & KDE \\
 \hline
1000  & 668.9 & \textbf{484.0} & 705.3  &  N/A &   \textbf{1.04s $\pm$ 1.12 ms} & 88s $\pm$344ms   \\
20000 & 767.7 & \textbf{535.4} &1100  &   N/A &   \textbf{20.7s $\pm$ 41.4 ms}& 2067s $\pm$78s   \\
$>$30000  & 834.1 &\textbf{569.8} & 1400  &   N/A &   \textbf{33.8s $\pm$ 24ms}& 2371s $\pm$35s   \\
\hline
\end{tabular}

\label{tab:mem_profile}
\end{table}

\begin{table}[!ht]
\caption{Execution run-time measured in seconds (s) $\pm$ standard deviation (std) for the $p$-value computation by  the different baselines for the ArtGAN DNN using the ArtWiki Dataset~\cite{artgan2018}. The sample size is reported as a \% of the total samples in the background data.
}
\centering
\begin{tabular}{@{}clll@{}}
\toprule
 & \multicolumn{3}{c}{$p$-value computation run-time (s) $\pm$ std}  \\ 
 \midrule
 \% Background samples & Empirical &   Proposed & KDE
 \\\midrule
20  & 0.024s $\pm$115$\mu$s &  \textbf{0.009s $\pm$10.4$\mu$s} & 32.9s $\pm$122ms   
\\
50 & 0.029s $\pm$49$\mu$s &  \textbf{0.009s $\pm$67.7$\mu$s} & 38.4s $\pm$65.7ms    
\\
100 & 0.033s $\pm$112$\mu$s &  \textbf{0.009s $\pm$13$\mu$s} &  44.2s $\pm$94.6ms    
\\
\bottomrule
\end{tabular}

\label{tab:p_val_compute_runtime}
\end{table}

\begin{figure}[!ht]
    \centering
    \resizebox{\linewidth}{!}{
    \begin{tabular}{@{}lccc@{}}
    \toprule
    Model Architecture & Empirical & Proposed\\
    \midrule
    \raisebox{60pt}{
    ArtGAN} 
    & {\includegraphics[width=0.45\textwidth]{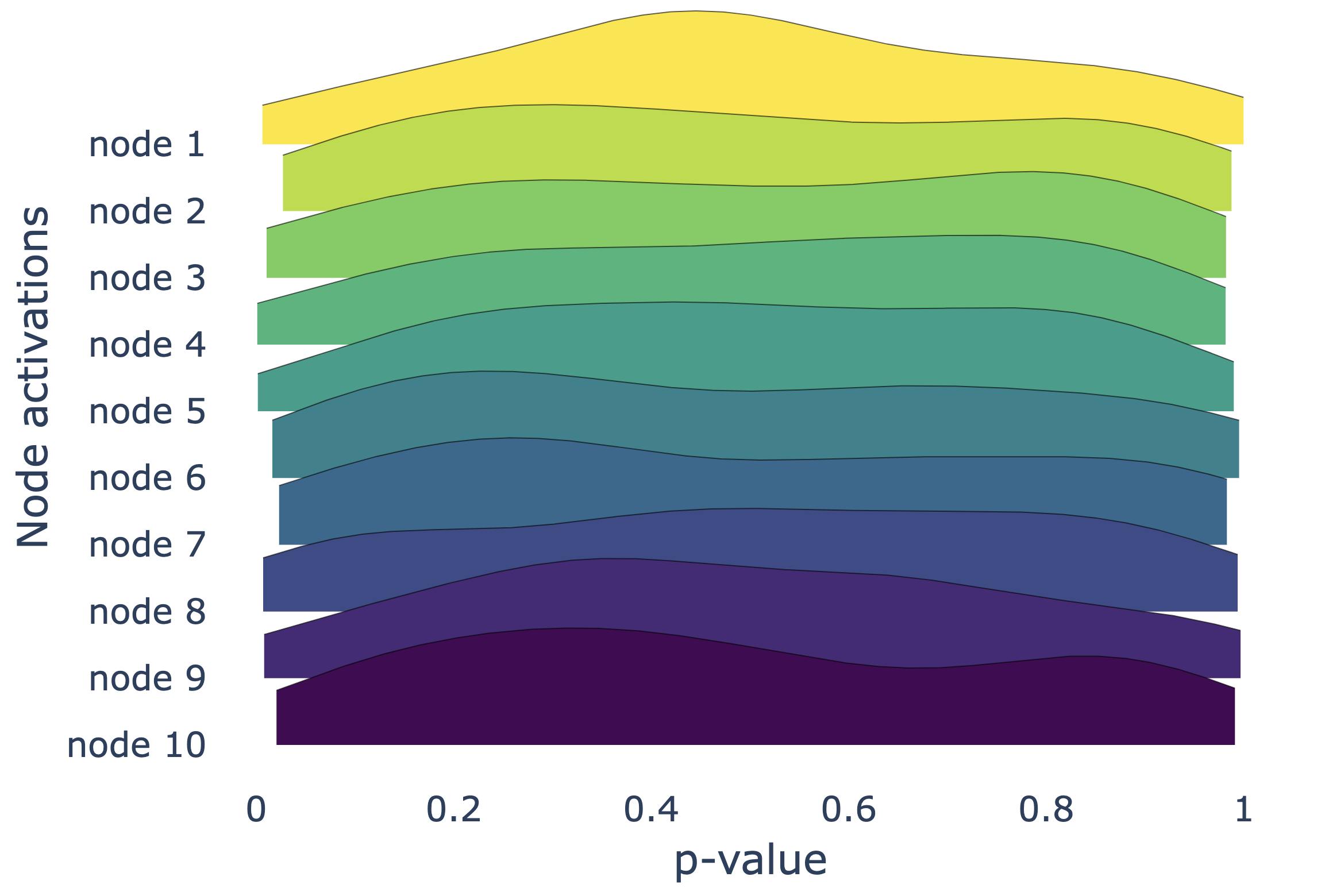}}
    & {\includegraphics[width=0.45\textwidth]{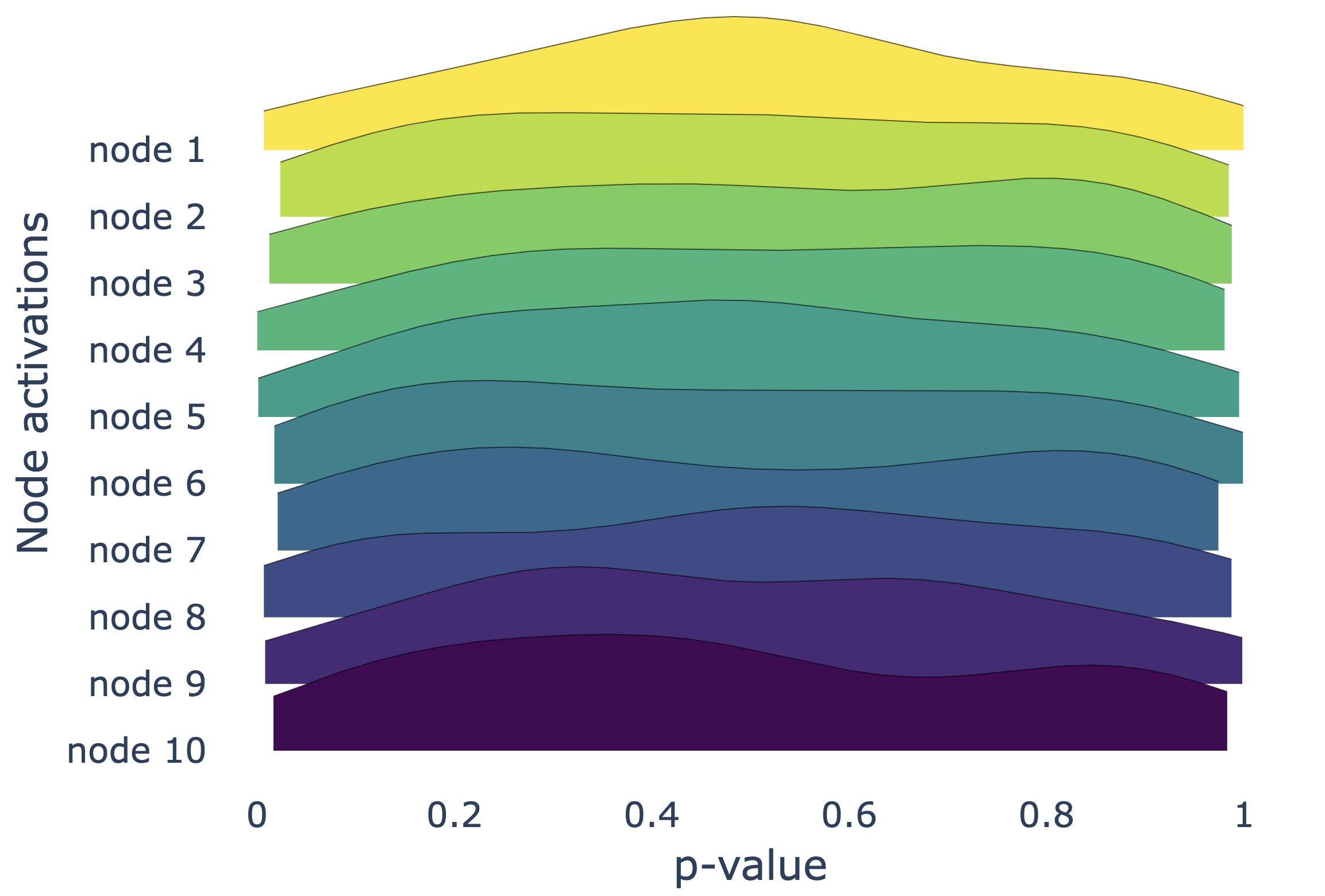}} \\
     & & $\overline{KS} = 0.06$, $p-value = 0.97$ \\
    \raisebox{60pt}{GCPN} 
    & {\includegraphics[width=0.45\textwidth]{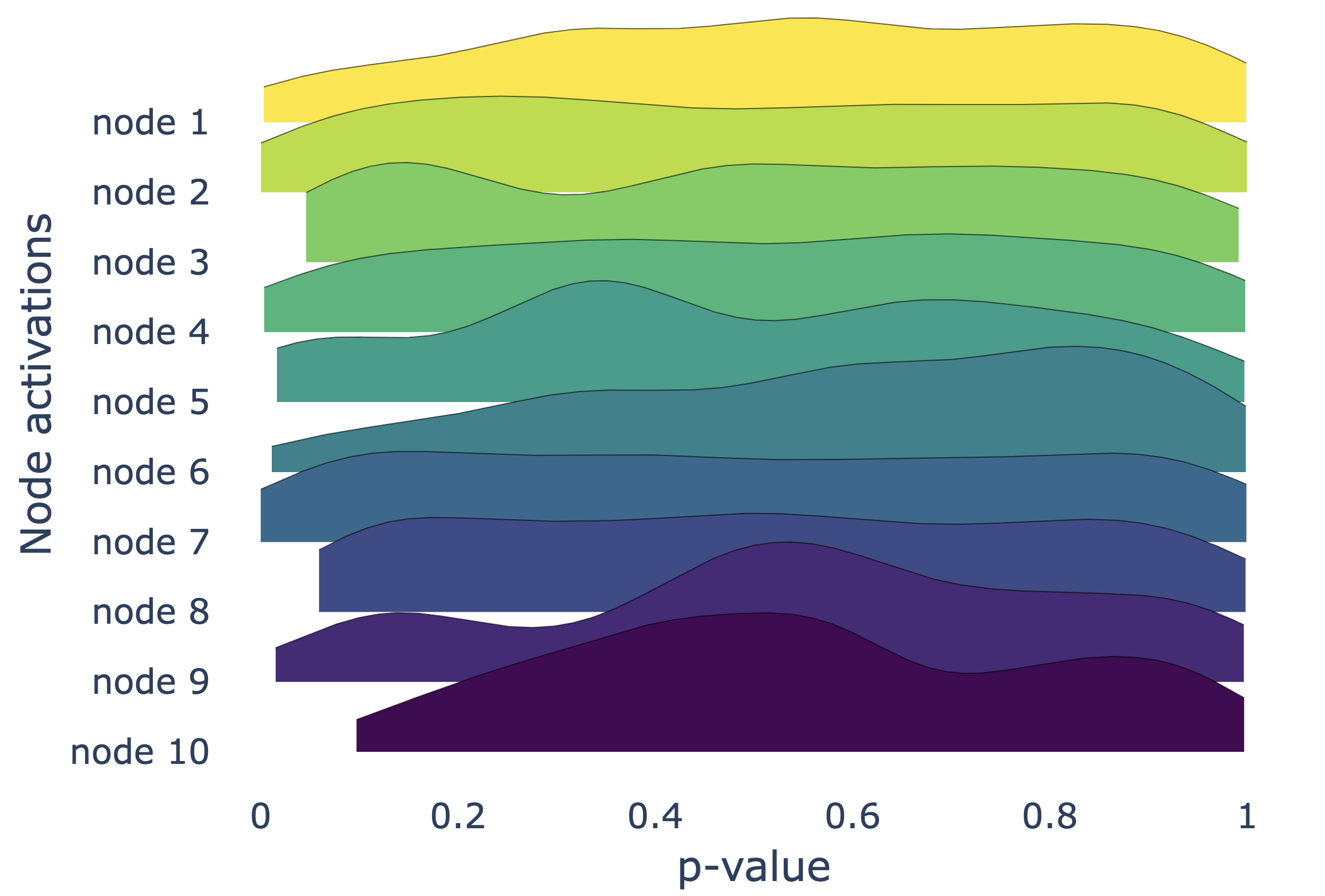}}
    &  {\includegraphics[width=0.45\textwidth]{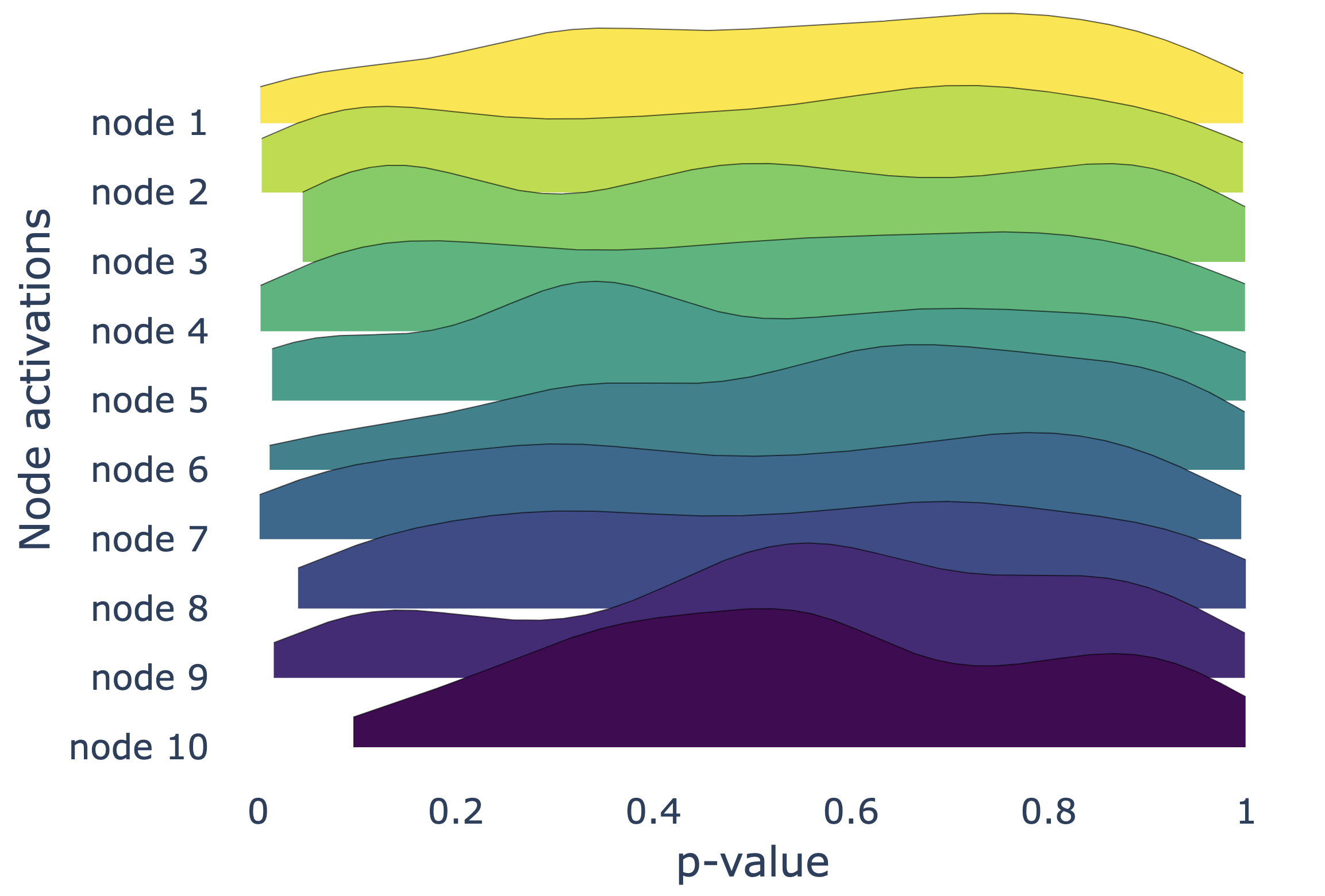}}\\
         &  & $\overline{KS} = 0.05$ , $p-value = 0.51$\\
    \raisebox{60pt}{CNN} 
    & {\includegraphics[width=0.45\textwidth]{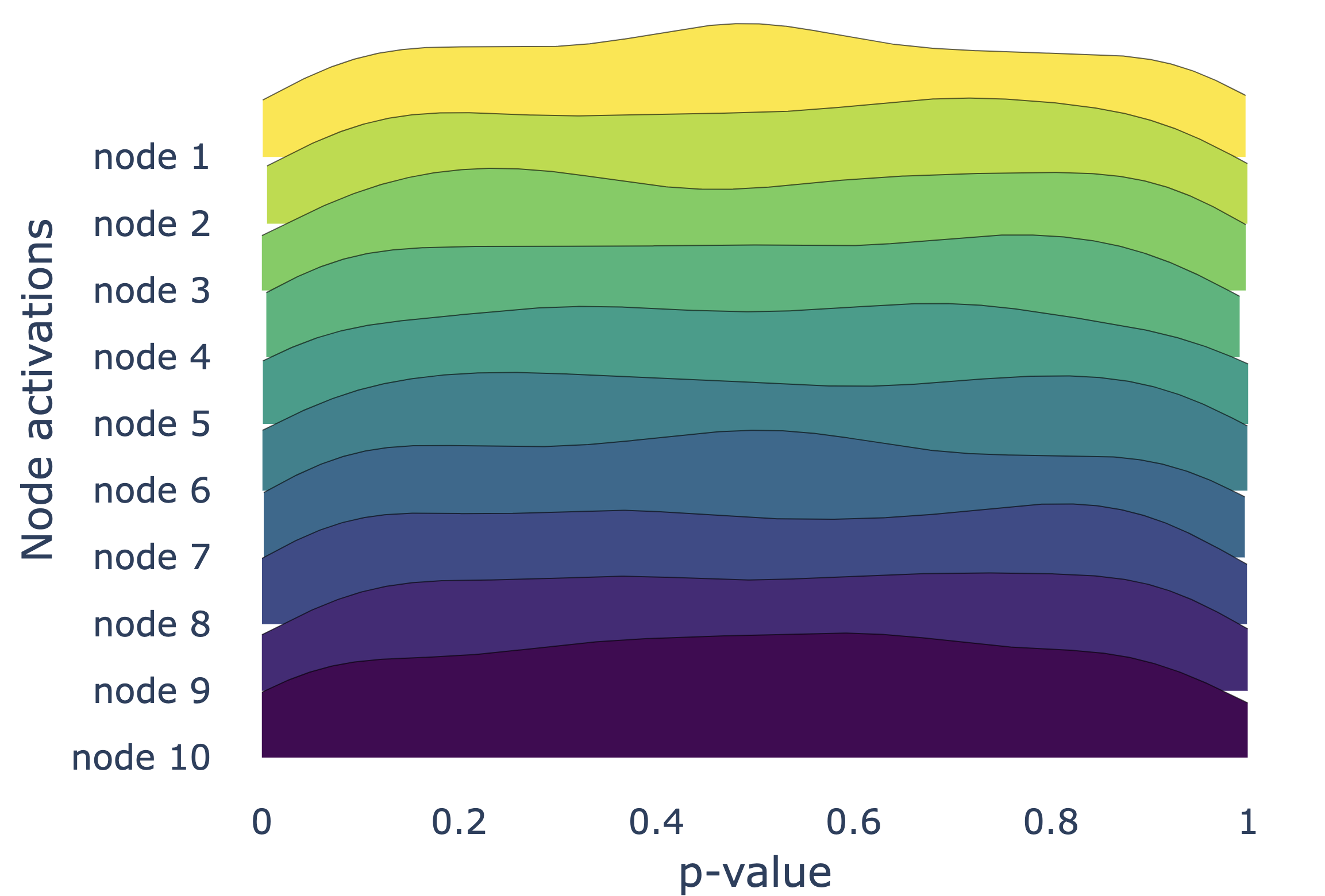}}
    & {\includegraphics[width=0.45\textwidth]{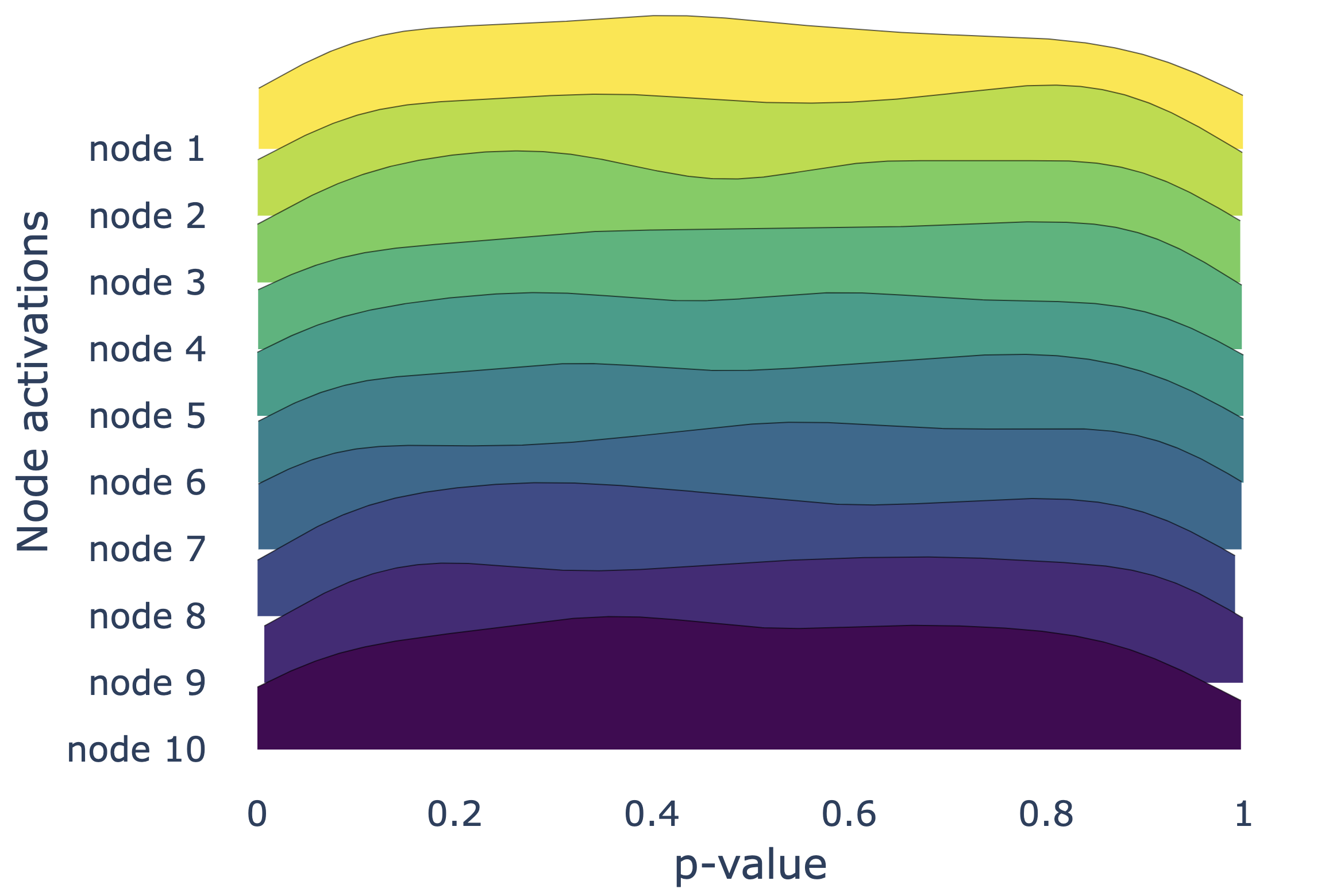}}\\
         & & $\overline{KS} = 0.04$ , $p-value = 0.7$ \\
    \bottomrule
    \end{tabular}
    }
    \caption{Ridgeplots comparing the empircal baselines with our proposed node-specific histogram $p$-values representation for the different DNNs and downstream tasks of creative~\cite{cintas2022towards} (ArtGAN), invalid graph~\cite{you2018graph} (GCPN) and adversarial detection~\cite{akinwande2020identifying} tasks (CNN). We can observe a visual similarity of distribution plots of the $p$-value representations of the node-specific histograms compared with the empirical baseline. This observed similarity is supported by the lower average $\overline{KS}$ statistic and the average $\overline{KS}$ $p$-value noted under each plot for the node-specific histograms.
    }
    \label{fig:p_val_ridgeplot}
\end{figure}

\section*{Experimental Setup}\label{experiments}
This section provides a detailed description of the experimental setup including the datasets used, the tasks performed, the histogram hyperparameter settings, and the evaluation metrics employed to assess the performance of our proposed approach.
We evaluated the performance of our proposed method under different downstream tasks and compared it against the empirical~\cite{mcfowland-fgss-2013,cintas2020detecting,feng-npss_graph-2014} and  KDE~\cite{feinman2017detecting,silverman1981using} baselines. 

\paragraph{Datasets \& tasks}
To demonstrate the generalizable and efficiency properties of the proposed framework, we evaluated the node-specific histogram $p$-value representations under multiple tasks and datasets that we will define below.

First, we look at the task of detecting creative artifacts~\cite{cintas2022towards}, where a Creative Generator variant of ArtGAN was used~\cite{das2021toward} for the WikiArt dataset~\cite{artgan2018}. The dataset contains paintings from $195$ different artists, and the dataset has $42129$ RGB images for training and $10628$ images for testing.
Second, we analyze the task of detecting invalid graph representation for molecule generation, using a Graph Convolutional Policy Network (GCPN)~\cite{you2018graph} as a base model and the ZINC dataset ~\cite{irwin2005zinc,sterling2015zinc}. This dataset contains $727842$ SMILES molecules. Lastly, we test our representation approach for detecting adversarial attacks in classical Convolutional Neural Networks (CNN) ~\cite{akinwande2020identifying} and Fast Gradient Signal Method (FGSM) attacks~\cite{goodfellow2015explaining} on the MNIST dataset~\cite{lecun1998mnist}. MNIST contains $60000$ images, and $10000$ images were extracted for this task. We selected the layer that was used in each of the cited papers as the more discriminative layer for a given task.

\paragraph{Extracting activations from DNNs}
Our approach requires constructing the histogram, $\Theta_{lj}$,  of the distribution of expected activations at each node of the layer, $\mathcal{L}$, of the pre-trained DNN, $M$. We  extract the expected activation distribution by passing forward the training data through $\mathcal{L}$ of the model $M$ and record the activations at each node. 
We create three sets of activation distributions. Namely, the background $A^{\mathcal{B}}_{lj}$ which is obtained by training $M$ on the background samples,$\mathcal{B}$, and clean and anomalous activations,$A^{\mathcal{T}_k}_{lj}$, which are obtained by training $M$ on test samples,$\mathcal{T}$. 
We assume that clean activations are drawn from the same distribution as the background samples, whereas the anomalous activations are from a noised sample of the training data passed through $M$. For the creative artifacts and adversarial attack detection tasks, we used $1000$ nodes in the activation space, whereas the invalid graph representation detection task had $256$ nodes.

\paragraph{Histograms hyper-parameter setup} 
Once we obtain the background activation distributions $A^{\mathcal{B}}_{lj}$, we create histograms for each node $j \in O_l$ of the selected discriminative layer. We remove modal activations that are greater than 10\% of the data, and manually add a unique bin for these groups of activations. We create histograms for activations outside the modal value with automatic binning, setting the maximum bin size to $10$. 

\paragraph{Performance evaluation}
The evaluation metrics used is the area under the receiver operating characteristic curve (AUROC) for detection power. To measure the anomalousness of a subset of node activations in a test sample, we use subset scanning~\cite{speakman_penalized, speakman2018subset}. This technique searches for the most anomalous subset of nodes within the inner layers of a DNN and calculates an anomalousness score for this subset, referred to as the 'subset score', using a scoring function.
For significance testing, we used the two-sample Kolmogorov-Smirnov-test, $\overline{KS}$,~\cite{massey1951kolmogorov} for estimating the distribution similarity of the $p$-value representations obtained by all baseline methods. Specifically, we average the KS-Statistic $\overline{KS}$ and $p$-value obtained from the test across all the nodes.

We also conducted benchmarks to profile the memory usage and run-time of our proposed approach and compared it to the other baseline methods.
We profile the execution run-time it takes to learn representations of the activation space offline, and the run-time for computing the $p$-value after obtaining these representations. These profiling tests were performed in a desktop machine (2.6 GHz 6-Core Intel Core i7, 32 GB 2400 MHz DDR4) under Linux 4.15.0-139-generic operating system. 

We conducted experiments to analyze the memory and execution time of the learning representation, we passed $500$ background samples through the pretrained ArtGAN while varying the number of nodes to $1000$, $20000$, and $32768$ (the total number of nodes in the selected layer) to obtain memory performance profiles. 


\section*{Results}\label{results}
\paragraph{Memory profile \& run-time benchmark}
In Table~\ref{tab:mem_profile}, we can see the memory profile of the baseline methods, and run-time for learning representations for the KDE~\cite{feinman2017detecting,silverman1981using} baseline and node-specific histograms approach for creating $p$-value representations of the ArtGAN. As observed, our method shows the best memory optimization when compared to the other methods. Our proposed approach reduces the memory used by about $30\%$, with a lower rise in memory usage as the number of nodes increases as compared to the other baselines. 
In terms of the learning representations run-time, the proposed method has a run-time of 33.8s, which is a huge improvement from the KDE baseline run-time of 2371s. Furthermore, in Table~\ref{tab:p_val_compute_runtime}, we see the proposed framework is not only four times as fast as the empirical method in computing the $p$-value representations, but also constant regardless of the number of samples in the background activations.

\paragraph{Downstream task performance}
Figure~\ref{fig:detection_pow_hist} reports the detection power through the AUROC score for the downstream tasks using the node-specific histogram $p$-values representation and the empirical $p$-values representation. We note that the detection power of our method's node-specific histograms $p$-values maintains the state-of-the-art results obtained by using the empirical $p$-values representation. The subset scores obtained from the scanning also show that the two baselines are comparable, indicating that the node-specific histogram approach is as good as the empirical baseline.

Lastly, in Figure~\ref{fig:p_val_ridgeplot}, we show a visual of the $p$-values distribution created by the empirical~\cite{mcfowland-fgss-2013,cintas2020detecting,feng-npss_graph-2014}, KDE~\cite{feinman2017detecting,silverman1981using}, and our proposed node-specific histogram method in the form of ridge plots for ten nodes of three DNNs for each downstream task. We observe that for the downstream tasks of creative artifacts~\cite{cintas2022towards}, adversarial \cite{akinwande2020identifying} and invalid graph representation detection~\cite{you2018graph}, the distributions of $p$-value representations created by the node-specific histogram approach are visually similar to those created by the empirical baseline. This is further supported by the low average $\overline{KS}$-test statistic of $0.06$, $0.05$ , $0.04$ for the DNNs in the proposed baseline. This $\overline{KS}$-test statistic indicates the maximum absolute difference between the distribution functions of the samples under test. Moreover, assuming a confidence level of 95\%, we see that the average $p$-value obtained from the $\overline{KS}$-test of $0.97$, $0.51$, and $0.7$ for the ArtGAN, GCPN, and CNN respectively, are not below the threshold of $0.05$. Therefore, we cannot reject the null hypothesis that the empirical $p$-value representations distribution and the node-specific histogram $p$-value representations distribution are similar. 

\section*{Conclusion \& Future Work}\label{conclusion}
We proposed a framework for creating representations of activations in DNNs based on node-specific histograms.  
We evaluate this approach by comparing it with two other baseline methods, namely, the empirical $p$-value representation \cite{mcfowland-fgss-2013,cintas2020detecting,feng-npss_graph-2014}and the Kernel Density Estimate $p$-value representation~\cite{feinman2017detecting,silverman1981using}, and validate it across different downstream tasks and datasets from multiple domains (chemistry, computational creativity, and adversarial machine learning).
The results show that our proposed approach is memory efficient and reduces the $p$-value computing time, while maintaining state-of-the-art performance in downstream tasks. Moreover, our approach extends the empirical $p$-values to $p$-value ranges which account for ties in likelihoods, thereby eliminating biases for higher $p$-values which could lead to more robust representations. Additionally, the use of representations of the activation space with node-specific histograms poses a potential advantage in privacy since the original data and full activations do not need to be shared. Although encouraging results in efficiency, the proposed approach faces some limitations. Currently, the proposed framework creates node-specific histograms for each node. This means that the run-time at inference increases linearly with the number of nodes. Future work would explore building histograms across subgroups of nodes across layers of DNN instead of individual nodes and using them to compute the $p$-value representations as well as other possible optimization strategies to further improve efficiency.



\setcitestyle{numbers}
\bibliographystyle{unsrtnat}
\bibliography{main}

\end{document}